\documentclass{article}
\usepackage{ijcai25}

\usepackage{times}
\usepackage{soul}
\usepackage{url}
\usepackage[hidelinks]{hyperref}
\usepackage[utf8]{inputenc}
\usepackage[small]{caption}
\usepackage{graphicx}
\usepackage{amsmath}
\usepackage{amsthm}
\usepackage{booktabs}
\usepackage{algorithm}
\usepackage{algorithmic}
\usepackage[switch]{lineno}

\usepackage{dblfloatfix}
\usepackage{adjustbox}
\usepackage{xcolor}
\usepackage{paralist}
\usepackage[shortlabels]{enumitem}
\usepackage{multirow}
\usepackage{cleveref}
\usepackage{adjustbox}
\usepackage{comment}
\usepackage{amssymb}
\usepackage{textcomp}
\usepackage{xcolor}
\usepackage{pifont}
\usepackage[most]{tcolorbox}

\newcommand{\cmark}{\textcolor{green}{\ding{51}}} % Green checkmark
\newcommand{\xmark}{\textcolor{red}{\ding{55}}}   % Red X

\title{Graph-based Molecular In-context Learning Grounded on Morgan Fingerprints}

\author{
Ali Al-Lawati
\and
Jason Lucas\and
Zhiwei Zhang\and
Prasenjit Mitra\and
Suhang Wang\\
\affiliations
The Pennsylvania State University\\
\emails
\{aha112, jsl5710, zbz5349, pmitra, szw494\}@psu.edu
}

\begin{document}

\maketitle

\begin{abstract}
%In-context learning (ICL) effectively conditions large language models (LLMs) for molecular tasks like property prediction and molecule captioning by incorporating carefully retrieved demonstration examples in the input prompt, avoiding the computational overhead of extensive fine-tuning. However, existing prompt retrieval methods for molecular tasks have relied on molecule feature similarity, such as Morgan fingerprint similarity, that fails to capture global molecular and atom-binding relationships, limiting their ability to represent the full complexity of molecular structures during inference. 

In-context learning (ICL) effectively conditions large language models (LLMs) for molecular tasks, such as property prediction and molecule captioning, by embedding carefully selected demonstration examples into the input prompt. This approach avoids the computational overhead of extensive pertaining and fine-tuning. However, current prompt retrieval methods for molecular tasks have relied on molecule feature similarity, such as Morgan fingerprints, which do not adequately capture the global molecular and atom-binding relationships. As a result, these methods fail to represent the full complexity of molecular structures during inference. Moreover, small-to-medium-sized LLMs, which offer simpler deployment requirements in specialized systems, have remained largely unexplored in the molecular ICL literature. % Moreover, the potential of small-to-medium-sized LLMs, which are deployable in specialized systems, has been largely unexplored in the molecular ICL literature.
To address these gaps, we propose a self-supervised learning technique, \texttt{GAMIC} (Graph-Aligned Molecular In-Context learning %\ali{Graph-Aligned Morgan fingerprint-based In-Context Learning})
, which aligns global molecular structures, represented by graph neural networks (GNNs), with textual captions (descriptions) while leveraging local feature similarity through Morgan fingerprints.  
In addition, we introduce a Maximum Marginal Relevance (MMR) based diversity heuristic during retrieval to optimize input prompt demonstration samples. Our experimental findings using diverse benchmark datasets show \texttt{GAMIC} outperforms simple Morgan-based ICL retrieval methods across all tasks by up to %across multiple small-to-medium-sized LLMs by up to 17.9\% on property prediction, 16.6\% on yield prediction, and 
45\%. %Our code is available at \url{https://anonymous.4open.science/r/mol-icl-AEF7}.
\end{abstract}

\section{Introduction}
Molecular representation and analysis field has significantly advanced towards specialized pre-trained language models like ChemBERTa~\cite{chithrananda2020chemberta}, and MolT5~\cite{edwards_translation_2022}. Through targeted pre-training and task-specific fine-tuning, researchers have achieved state-of-the-art (SOTA) results in molecular property prediction~\cite{tong2022blood,liu2023molxpt}, molecule captioning~\cite{he_g-retriever_2024,jiang2024enhancing}, and yield prediction~\cite{guo_what_2023,shi2024prediction}.%\suhang{the citation format of \cite{tong2022blood} doesn't seem to be correct. Please check the bib file. You should not add et al in the bib file. Put the full author list. There are similar issues for some other citations}  %\suhang{the citation format of \cite{guo_what_2023} is also incorrect}. 

Nonetheless, recent developments in large language models (LLMs) have demonstrated remarkable capabilities in prediction tasks through in-context learning (ICL) \cite{NEURIPS2020_1457c0d6}, potentially offering a more efficient alternative to the computationally expensive pre-train and fine-tune paradigm.  %\suhang{One sentence introducing ICL, e.g., Generally, given a target molecular for analysis using LLM, ICL retrives xxx to xxx.} 
Generally, given a target molecular for molecule captioning or property prediction using LLM, ICL retrieves similar molecules with their captions or properties, and uses these retrieved examples in the prompt as demonstration~\cite{guo_what_2023,li_empowering_2024}, which provides important information to guide LLMs to give more accurate predictions. While this approach can enhance prediction accuracy, its effectiveness heavily depends on both the relevance and diversity of the demonstration samples used to guide the LLM~\cite{das-etal-2021-case}. 
Despite this, the effectiveness of ICL remains underexplored in molecular tasks, particularly for small to medium-sized LLMs ($\raisebox{0.2ex}{$\scriptstyle<$}$ 10B)~\cite{wang2024comprehensive} such as Mistral-7B~\cite{jiang2023mistral}.

Recently, researchers have introduced Morgan fingerprint-based methods, such as \textit{Scaffold}~\cite{lim2020scaffold}, for ICL demonstration selection~\cite{guo_what_2023}, which utilizes the similarity of the Morgan fingerprint between the test sample and the demonstration pool. Although Scaffold outperforms random selection, its reliance on Morgan fingerprints only constrains its ability to retrieve structurally similar samples for ICL, as Morgan fingerprints cannot fully encode the complex binding relationships that are better represented by molecular graphs~\cite{jin2018learning}.
% However, while Scaffold has shown improvement over random selection, Morgan fingerprints do not encode the full complexity and binding relationships of molecules that are more accurately represented as a graph~\cite{jin2018learning}, limiting their ability to retrieve structurally similar molecules for better ICL. %\suhang{can we add one or two sentences explaining why capturing the graph structure is important for molecular analysis} \ali{
Thus, capturing the graph structure is crucial for molecular analysis because it preserves atoms' spatial and connectivity information. This detailed representation is particularly important for molecular similarity retrieval, where subtle structural variations can significantly impact chemical behavior. This raises a natural question: \textbf{Can we combine the graph representation of the molecule with the Morgan fingerprint to further enhance ICL effectiveness by capturing both local properties (captured in the Morgan fingerprint) and global molecular structures (represented by a graph)?}
%describe it from two perspectives: graph retrieval + MMR. Make it two sections
%\suhang{the logic for the following few paragraphs should be like this: (i) However, there are several changes to answer this question. List and explain the challenges; (ii) To address these challenges, we proposes a novel framework xxx. Briefly introduce the framework and how it addresses the challenges.} \ali{[Connecting sentence for GNN].

To explore this possibility, a leading approach is to leverage Graph Neural Networks (GNNs)~\cite{scarselli2008graph}, which are the SOTA method for processing molecular graph structures~\cite{wang2022molecular}. However, applying GNNs in molecular similarity retrieval presents several challenges. In particular, \begin{inparaenum}[(i)]
    \item GNN encoding struggles to convert complex discrete molecular structures into continuous latent spaces while preserving chemical validity~\cite{edwards2021text2mol}, i.e. \textit{complexity challenge}; %\suhang{it's unclear to me how the proposed framework addresses this challenge. could you explain in the paragraph below and also explain this in the methodology part (so we can explain why we design in this way in the methodology part)}
    \item GNN learning on multimodal datasets, such as PubChem~\cite{pubchem}, is susceptible to information loss due to the significant gap between graph and text representations~\cite{song2024towards}, i.e. \textit{modality gap}; %\suhang{it's unclear to me how the proposed framework addresses this challenge. COuld you explain in the paragraph below and also explain this in the methodology part }
    \item public datasets describe molecules in various ways, ranging from concise single-sentence descriptions to detailed multi-sentence explanations that capture very specific details,
    % molecules in public datasets are described in diverse and non-standarized methods
    ~\cite{liu_molca_2023}, i.e. \textit{dataset limitations}, %\suhang{I cannot get dataset limitation. What does diverse non-standardized methods means? Why this is a challenge to our ICL? I cannot get it. I would just delete this challenge if we can not explain it well}%, 
    which further exacerbates the modality gap.
\end{inparaenum}

%This underscores the need for a unified approach combining reasoning capabilities with effective representation learning to bridge the gap between different molecular tasks while taking advantage of the computational simplicity of the ICL paradigm. 

To address these challenges, %\suhang{it is not clear to me what are the chalelnges}, 
we propose \texttt{GAMIC} (\textbf{G}raph-\textbf{A}ligned \textbf{M}olecular \textbf{I}n-Context learning), a novel ICL method that leverages the inherent graph structure of molecules and their local molecular features for multimodal graph-language training. In particular, \texttt{GAMIC} processes the molecular representation using a hierarchical graph encoder and aligns the latent representation with their scientifically-aware (e.g. SciBERT~\cite{Beltagy2019SciBERT}) embedded textual descriptions using a sampling method based on Morgan fingerprint similarity. Incorporating Morgan fingerprints as a preliminary step to select alignment pairs helps narrow the \textit{modality gap} by providing a robust and interpretable measure of local molecular similarity during multimodal alignment training. In addition, using scientifically-aware textual embedding enriches the latent space representation of the encoded graph post-alignment, mitigating the \textit{complexity challenge}. Finally, by expanding the pool of potential textual representations grounded on Morgan fingerprints, \texttt{GAMIC} provides a more robust solution to address \textit{dataset limitations}. Moreover, to further enhance ICL retrieval, we introduce a novel diversity-aware sample selection method using Maximum Marginal Relevance (MMR) to maximize the information provided in the input prompt.
%\suhang{we generally do not mention the model architecture in the introduction} In order to accomplish this, 
%In particular, \texttt{GAMIC} implements a hierarchical graph capture through a specialized graph encoder with a graph-text contrastive learning method to align graphical representations with textual descriptions conditioned on Morgan fingerprints. During retrieval, we utilize for diversity-aware sample selection.  \suhang{merge this paragraph with the above paragraph to introduce high-level idea of the framework and how it addresses the challenges}. 
Our key contributions are:
\begin{itemize}

    \item A novel multimodal ICL method for molecular tasks using graph molecular features grounded on Morgan fingerprint-based sampling.
    \item An MMR-based demonstration selection heuristic to enhance sample diversity.
    \item Comprehensive experimental evaluation comparing our approach with existing methods using three medium-size general-purpose LLMs.
    %\item Insights into the ICL capabilities of smaller LLMs in molecular tasks using different ICL approaches, including \texttt{GAMIC}.
\end{itemize}

\begin{figure*}[t]
    \centering
        \includegraphics[width=.82\textwidth]{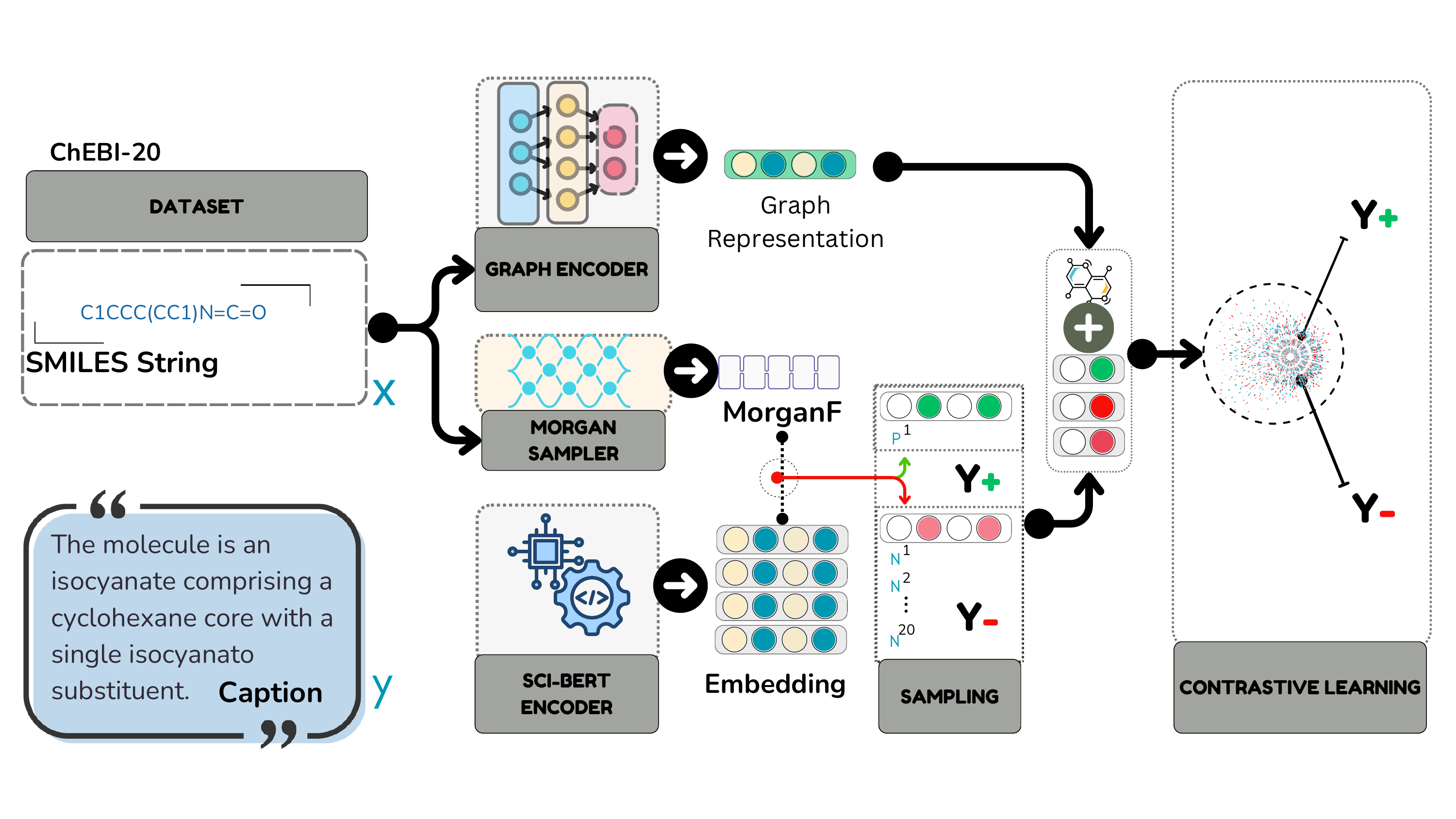} % Adjust the `trim` values as needed

    \vskip -2em
    \caption{Overview of \texttt{GAMIC} Graph Projector}% \suhang{the figure is a little bit confusing: (i) In the figure, it seems that the graph encoder gives a graph. However, in fact, it gives the graph representation. (ii)  why the sci-bert encoder looks like a graph???}}
    \label{fig:graph}
   % \vspace{-10pt}
\end{figure*}

\section{Related Work}

\subsection{Molecular Representation Learning}
Traditional molecular modeling approaches have predominantly relied on specialized architectures that directly operate on molecular structures for tasks such as property prediction~\cite{guo2021few,stark20223d}, molecule generation~\cite{gong2024text,kim2024data}, and reaction prediction~\cite{liu2024reactxt}. With the advent of the transformer architecture~\cite{vaswani2017attention}, the field has witnessed a shift towards representation learning through pre-training and fine-tuning paradigms. Early transformer-based approaches focused on learning from SMILES~\cite{weininger1988smiles} string representations. For example, MolBERT~\cite{li2021mol} adapted the BERT~\cite{Devlin2019BERTPO} architecture to recognize different SMILES~ string representations of compounds, while ChemBERTa~\cite{chithrananda2020chemberta} employed masked language modeling (MLM) on text-SMILES datasets. More recent approaches have explored richer molecular representations and transfer learning. MolT5~\cite{edwards_translation_2022} finetunes a pre-triend T5 language model for moleculecular translation. %\suhang{MolT5~\cite{edwards_translation_2022} finetunes a pre-triend general-purpose language models XXX for xxx}, 
MolCA~\cite{liu_molca_2023} introduced a cross-model projector to effectively fine-tune LLMs on select downstream tasks, %\suhang{then why not use MolCA for molecular graph representation learning in ICL??? You might want to either delete this or write it in a way that will not lead reviewers to ask the question I raised, e.g., explicitly write done fine-tuning LLM part}
while 3D MolM enhanced existing datasets by incorporating 3D conformational information generated using GPT-3.5.

Despite their effectiveness in molecular representation learning and analysis, these pre-training and fine-tuning approaches face the following limitations:
\begin{inparaenum}[(a)]
    \item requirements for significant computational resources during pre-training,
    \item need for task-specific fine-tuning and separate training for each task, and
    \item limited flexibility in adapting to new molecular tasks.
\end{inparaenum}
\subsection{In-Context Learning for Molecular Tasks}
ICL has emerged as a promising alternative for the pre-train/fine-tune paradigm, enabling general-purpose language models to perform various tasks through demonstration-based prompting. Instead of fine-tuning, ICL provides demonstrations in the prompt, which allows the LLM to learn from them and generate more accurate responses.   
Despite the effectiveness of ICLs in various applications~\cite{dong2022survey}, the work on ICL for molecular tasks is still in its early stage and there are very few works~\cite{li_empowering_2024,guo_what_2023}. %~\suhang{can we add this: The work on in-context learning for molecular tasks is still in its early stage and there are very few works~(cite these works)} 
Recently, MoleReGPT~\cite{li_empowering_2024} introduced dual approaches for molecular tasks. For molecular captioning, MoleReGPT utilizes Morgan fingerprint similarity, i.e., Scaffold, which compares the presence of specific substructures encoded in the Morgan fingerprint vector. %\suhang{and encode them in a vector???} encoded in a vector, i.e. Morgan fingerprint. %, and for molecule generation, it uses BM25 %\suhang{can you introduce more details of Morgan similarity and probably Scaffold since we used it as a baselines???} 
Guo et al.~\cite{guo_what_2023} established a benchmark across eight molecular tasks, evaluating various LLMs using random and scaffold-based sample selection. However, existing ICL approaches for molecular tasks have several limitations:
\begin{inparaenum}[(a)]
    \item insufficient capture of bond connectivity and atomic features present in molecular graphs,
    \item limited exploration of graph-aware contrastive learning for demonstration selection and
    \item primarily focus on large and commercial language models, such as GPT4.
\end{inparaenum}

While GNNs have demonstrated promise in capturing molecular structure in fine-tuned model such as  MolCA~\cite{liu_molca_2023}, their potential for enhancing ICL demonstration selection remains underexplored. Our work addresses this gap by introducing \texttt{GAMIC}, the first approach to leverage Morgan-based graph alignment, achieving SOTA performance on benchmark molecular ICL tasks. This novel direction addresses the limitations of existing methods while maintaining computational efficiency central to the ICL paradigm.

\section{Methodology}
\begin{figure}[t]
    \centering
        \includegraphics[width=.45\textwidth]{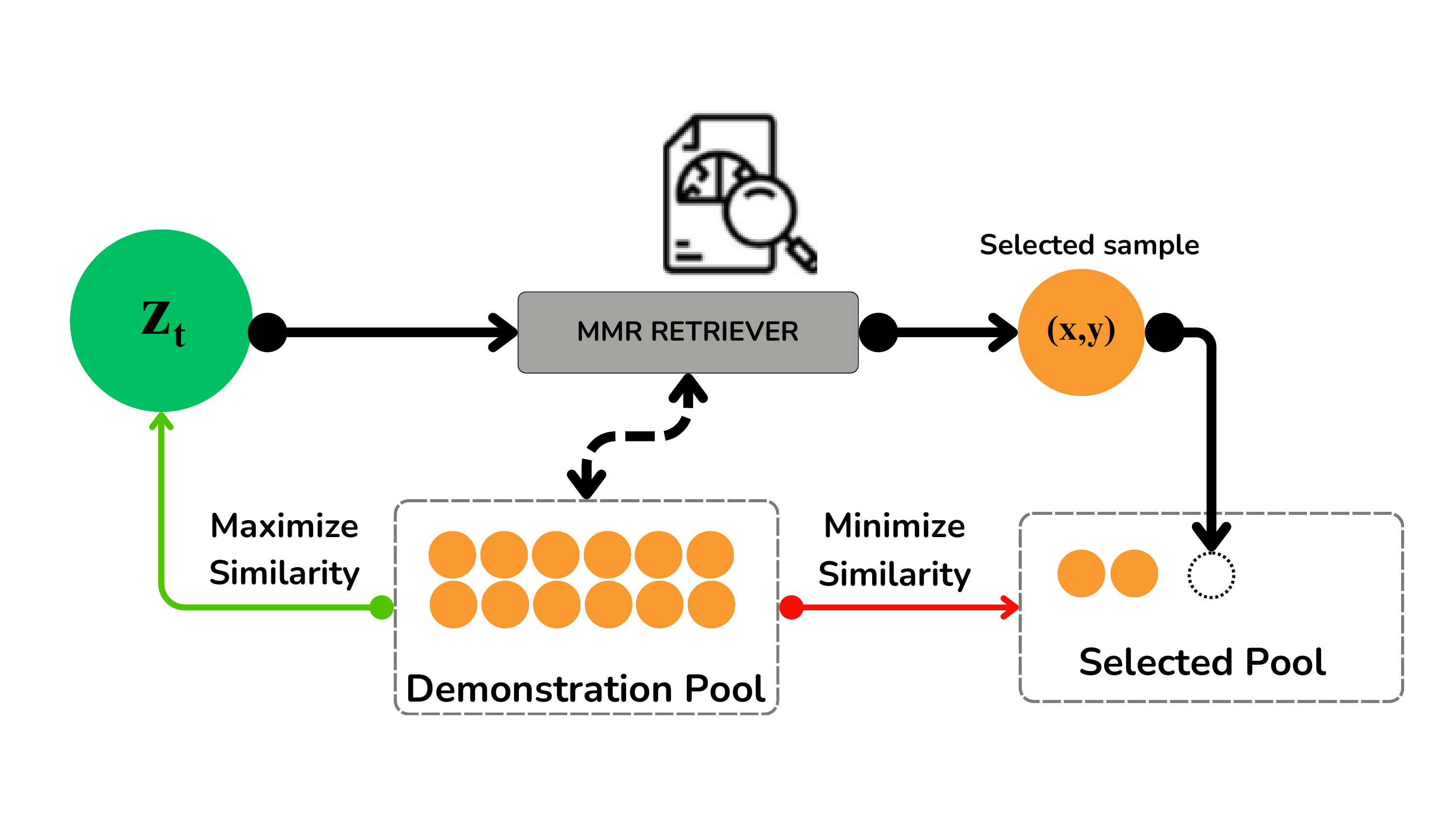} % Adjust the `trim` values as needed

    \vskip -2em
    \caption{MMR-based Sample Selector}% \suhang{this figure looks good. I think we might need to have another figure showing the ICL process with MMR}}
    \label{fig:mmr}
   % \vspace{-10pt}
\end{figure}

In this section, we will first give the problem definition, then the overview of the proposed \texttt{GAMIC}  followed by the details of \texttt{GAMIC}.

\subsection{Problem Setup}
Given a training set $\mathcal{T} = {(x_i, y_i)}_{i=0}^n$ of molecule-value pairs with $x_i$ as a SMILES string and $y_i$ as the corresponding value, we aim to learn a graph retriever $R$, such that given a test molecule $x_t$, the \texttt{GAMIC} retriever can retrieve relevant and diverse demonstration $P_t = R(x_t, \mathcal{T})$ from a demonstration pool, which will be concatenated with $x_t$ and prompt as input to an LLM $\mathcal{M}$ for molecular analysis. The objective of the \texttt{GAMIC} retriever is to select $P_t$, such that $\mathcal{M}(P_t; x_t)$ will yield $y_t'$, that maximizes $\mathcal{D}(y_t, y_t')$, where $\mathcal{D}$ is a similarity metric (e.g., BLEU score~\cite{papineni2002bleu}) and `;' represents concatenation.
%Given a pre-trained language model $\mathcal{M}$ and a training set $T = {(x_i, y_i)}_{i=0}^n$ of molecule-caption pairs, we select $k$ demonstrations $P \subset T$ to construct a prompt. For a test molecule $x_t$, we concatenate $P$ with $x_t$ as input to $\mathcal{M}$ to generate output $\hat{y}$, aiming to maximize semantic similarity $f$ (e.g., BLEU score) between $\hat{y}$ and ground truth $y_t$.
%\suhang{I think the problem definition is not correct: (i) $y_i$ is the caption, doesn't necessary to be the label that we want to predict; (ii) the demonstration $P$ is sample dependent, while currently the description is more like $P$ is independent of samples. I think it should be something like this: Given a training set $T = {(x_i, c_i)}_{i=0}^n$ of molecule-caption pairs with $x_i$ as a molecule graph and $y_i$ as the corresponding caption, we aim to learn a graph retriever $R$, such that given a target molecular graph $x_t$, the graph retriever can retrieve relevant and diverse demonstration $P_t = R(x_t; P)$ from a demonstration pool, which will be concatenated with $x_t$ and prompt as input to an LLM for molecular analysis.}

%The model is prompted using the selected demonstrations concatenated with the test molecule, i.e., $\mathcal{M}(P | x_t)$, where $|$ represents concatenation. The semantic similarity is quantified by a metric $f$, such as the BLEU score~\cite{papineni2002bleu}.

\begin{table*}[t]
\centering
\caption{Overview of tasks, datasets, and evaluation metrics}
\label{tab:dataset_stats}
\vskip -1em
\begin{adjustbox}{width=.9\textwidth}
\begin{tabular}{lllrrc}
\toprule
\textbf{Task} & \textbf{Task Class} & \textbf{Dataset} & \textbf{Test Size} & \textbf{ICL Pool Size} & \textbf{Ev. Metrics} \\
\midrule
\multirow{2}{*}{Molecule Captioning} & \multirow{2}{*}{Molecular Explaining} 
& ChEBI-20 & 3300 & 26407 & BLEU, ROUGE,\\ 
& & PubChem & 2000 & 12000 & METEOR\\
\midrule
\multirow{2}{*}{Yield Prediction} & \multirow{2}{*}{Molecular Reasoning} 
& Suzuki-Miyaura & 576 & 4608 & \multirow{2}{*}{F1-score/StDev}\\
&& Buchwald-Hartwig & 396 & 3163\\
\midrule
\multirow{5}{*}{Property Prediction} & \multirow{5}{*}{Molecular Understanding} 
& BBBP & 204 & 1631  & \multirow{5}{*}{F1-score/StDev}\\
& & BACE & 152 & 1209 \\
& & HIV & 4113 & 32901 \\
& & Tox21 & 784 & 1184\\
& & ClinTox & 148 & 6264\\
\bottomrule
\end{tabular}
\end{adjustbox}
\end{table*}

\subsection{Overview of Model Architecture}

% \suhang{TODO: need to be significantly revised}
The proposed framework, \texttt{GAMIC}, is composed of two parts, i.e., (i) Graph Projection (see \Cref{fig:graph}), which aims to learn graph representation of a molecualr graph that captures both bond connectivity and atomic features for demonstration retrieval; and (ii) MMR-based sample selection (see \Cref{fig:mmr}), which aims to select similar and diverse demonstrations to improve the performance of an LLM. 
% \begin{itemize}
%     \item Graph Projector (see \Cref{fig:graph})
%     \item MMR-based sample selector (see \Cref{fig:mmr})
% \end{itemize}
%\suhang{Can we describe our framework as composed of two parts: 1. Graph-aware Retriever (you can come up with you own name); and 2. MMR based selection.}
%\Cref{fig:graph} illustrates the high-level training procedure for \texttt{GAMIC}. 
Specifically, the graph projection adopts a \textbf{Graph Encoder} to learn graph representation of molecular graphs constructed from SMILE Strings. To train the graph encoder, it adopts contrastive learning and utilizes a \textbf{Morgan Sampler} to find positive and negative alignment candidates for contrastive learning. The encoder is trained to learn graph representation that align with positive textual captions encoded using the \textbf{SciBERT Encoder} using Contrastive Learning, as depicted in \cref{fig:graph} . %A MMR-based sample selector is used during ICL demonstration retrieval. Next, we give details of each component. 
During the ICL demonstration retrieval process, MMR-based Sample Selector retrieves informative and diverse examples.  
Next, we describe each component of \texttt{GAMIC} in more detail.

\subsection{Graph Projection} 
\subsubsection{Graph Encoder} 
To sufficiently capture the bond connectivity and atomic features present in molecular graphs, given a training set of $(x, y)$ pairs, where $x$ is the SMILES string, and $y$ is the natural language description, i.e. caption, we construct a molecular graph for each SMILES string ($x$)%\suhang{$x$???})
: \( G = (\mathbb{V}, \mathbb{E}) \) with atoms as nodes \( \mathbb{V} = \{v_1, \ldots, v_N\} \) and bonds as edges \( \mathbb{E} \). With the molecular graph, we use a two-layer Graph Attention Network (GAT) \cite{gat} to learn node representation as
%We represent \(G\) using three matrices: adjacency matrix \( \mathbf{A} \in \mathbb{R}^{N \times N} \) encoding connectivity, node feature matrix \( \mathbf{X} \in \mathbb{R}^{N \times F} \) containing atom properties, and edge feature matrix \( \mathbf{E} \in \mathbb{R}^{M \times D} \) encoding bond attributes. We denote the test graph as \(G_t\) and selected demonstrations as \(G_1, \ldots, G_m\).
%\subsubsection{Graph Projection}
%We encode each graph using a two-layer Graph Attention Network (GAT):
\begin{equation}
\mathbf{H} = \text{GAT}(\mathbf{X}, \mathbf{A}, \mathbf{E}; \theta_{GAT}),
\end{equation}
where \( \mathbf{A} \), \( \mathbf{X} \), and \( \mathbf{E}\) are the adjacency matrix, node features, and edge features, respectively. Next, we apply a pooling on top on node representation followed by a MLP to obtain the final graph embedding, \(\mathbf{z}\), as follows
\begin{equation}
\mathbf{z} = \text{MLP}( \text{MeanPool}\left(\mathbf{H} \right), \mathbf{w}^{(0)}),
\end{equation}
%\suhang{you do not need to specify it as a two-layer GAT. You just simply use $\mathbf{H} = GAT(\mathbf{X}, \mathbf{X}, \mathbf{E}; \theta_{GAT})$. The two-layer GAT can be explained in the detailed implementation in the experiment.} 
where \(\mathbf{w}^{(0)}\), is a learnable weights, and \(\sigma\) is ReLU activation.

\subsubsection{Morgan Sampler}%Similarity Sampling Based Contrastive Learning} %\suhang{do not call this retriever, which will cause confusion to our retriever. Instead, use Morgan Simialrity Based Contrastive Learning} 
%\suhang{need to explain why not simply treat a graph with its corresponding caption as positive, what's the advantage of using Morgan Similarity Retriever???} %\suhang{The logic of the following paragraphs is strange. We need to first explain the higher level idea, e.g., In order to train the retriever that are well at xxx, we propose to adopt contrastive learning. One way is to treat xxx as positive. However, it will cause xxx issue. Therefore, we propose to adopt Morgan Similarity based xxx. }

In order to train the graph projector to align the final graph embedding with the captions, we propose adopting contrastive learning. Our preliminary testing showed that multimodal contrastive learning significantly outperforms other graph-based approaches such as graph autoencoder, or traditional graph-based contrastive methods. %\suhang{it is unclear why we want to align the final embedding with the captions? Why not use autoencoder that reconstructs Adjacency matrix? Why not use traditional contrastive learning that only uses graphs? What's the advantage? I fell it is better to add some explanations here} 
Hence, for each graph, we treat the corresponding caption as positive, and randomly sample a negative pair(s) from the dataset. However, this may cause information loss due to the modality gap, as discussed above. In addition, dataset limitations, characterized by varying number of sentences in the captions or the type of details described, may hinder a robust alignment. 

Therefore, we propose adopting Morgan fingerprint-based sampling ($\mathcal{R}_m$) to expand the sets of positive and negative caption pairs for alignment by including molecules with similar Morgan fingerprints. For each training sample, $x_i$, $\mathcal{R}_m(x_i)$ returns $\mathcal{Y}_{i}^{+}$, %\suhang{let's use $\mathcal{Y}_{i}^{+}$ as set for $x_i$} 
a set of positive samples, and $\mathcal{Y}_{i}^{-}$, a set of negative samples, based on Morgan fingerprint similarity between $x_i$ and the training set at each epoch.

\subsubsection{SciBERT Encoder} To align the graph representation with texts, we need to get text representation first. we adopt SciBERT~\cite{Beltagy2019SciBERT} %~\suhang{cite SciBERT~\ali{i cited it above, should I cite it again here?}} 
as the text encoder.
% SciBERT is a domain-specific model trained on a large corpus of scientific texts that covers scientific terminology used in molecular captions better than general-purpose models~\cite{li2024molm}, such as BERT:
SciBERT is a domain-specific model trained on a large corpus of scientific texts, providing better coverage of scientific terminology in molecular captions compared to general-purpose models~\cite{li2024molm} like BERT.  Specifically, for each caption $y \in \{\mathcal{Y}^+,\mathcal{Y}^-\}$, we obtain a fixed-size embedding using SciBERT as:
\begin{equation}
    y_{emb} = \text{SciBERT}(y)
\end{equation}

\subsubsection{Contrastive Learning}
% To address the limitations in existing work, including the limited exploration of graph-aware contrastive learning for demonstration selection and the lack of robust diversity-aware selection heuristics, 
% we utilize contrastive loss~\cite{oord2018representation} between graph embeddings and their corresponding text and fingerprint representations:
Existing work on ICL has been limited by a lack of focus on graph-aware contrastive learning. To address this limitation, we propose utilizing a contrastive loss~\cite{oord2018representation} that aligns graph embeddings with their corresponding text representations. The contrastive loss is formulated as:
\begin{equation}
 \mathcal{L} = \text{NCE}(\mathbf{z}, \mathcal{Y}_{emb}^+, \mathcal{Y}_{emb}^-),   
\end{equation}
%where positive pairs (\(\mathbf{T}^+, \mathbf{F}^+\)) are from the same molecule as \(\mathbf{Z}\), and negative pairs (\(\mathbf{T}^-, \mathbf{F}^-\)) are from different molecules. 
where the Noise Contrastive Estimation (NCE) function is defined as:
%\begin{adjustbox}{width=.5\textwidth}
\begin{equation*}
\scalebox{0.9}{$\text{NCE}(\mathbf{z}, \mathcal{Y}^+, \mathcal{Y}^-) = -\frac{1}{N} \sum_{i=1}^{N} \log \left( \frac{\exp\left( \mathbf{z}_i \cdot y_i^+ / \tau \right)}{\exp\left( \mathbf{z}_i \cdot y_i^+ / \tau \right) + \sum_{j=1}^{K} \exp\left( \mathbf{z}_i \cdot y_{ij}^- / \tau \right)} \right)$}
\end{equation*}

%\suhang{do not give the general NCE loss function, give ours, i.e., $\text{NCE}(\mathbf{z}, Y_{emb}^+, Y_{emb}^-)$}
where $\tau$ is a temperature parameter that controls the sharpness of the similarity distribution, and subscript ($_{emb})$ is omitted for all $y$ for readability.

\subsection{MMR-based Sample Selector}
% For $(x_t,y_t)$, we select $k$-samples $(x_1, y_1), \ldots, $ $(x_k,y_k)$ in the following way:
During retrieval, we ensure both relevance and diversity in demonstration selection by employing a Maximal Marginal Relevance (MMR)-based selection strategy. For a given test sample $(x_t, y_t)$, we select $k$ demonstrations $(x_1, y_1), \ldots, (x_k, y_k)$ by solving the following optimization iteratively: %\suhang{please make the notation consistent, i.e., use $\mathbf{z}_i$ as the molecular graph representation}
\begin{equation}
     \min_{\mathbf{z}\in P} \| \mathbf{z}_i - \mathbf{z}_t \| + \lambda \sum_{j=1}^{i-1} \max \|\mathbf{z}_i - \mathbf{z}_j \| \\ 
\quad  \text{for} \quad  i \in 1,\ldots,k
\end{equation}
%\suhang{I think it should be something like
%\begin{equation}
%     \min_{\suhang{\mathbf{z}\in P}} \| \mathbf{z} - \mathbf{z}_t \| + \lambda \sum_{j=1}^{i-1} \max \|\mathbf{z}_i - \mathbf{z}_j \| \\ 
%\quad  \text{for} \quad  i \in %1,\ldots,k
%\end{equation}
where $P$ is the set of possible demonstrations and $\mathbf{z}$ is the latent representation of $x$, and $\lambda$ is a hyperparameter that balances relevance to the test sample (minimizing $\|\mathbf{z}_i - \mathbf{z}_t\|$) with diversity among the selected demonstrations (maximizing $\|\mathbf{z}_i - \mathbf{z}_j\|$). This approach ensures that selected demonstrations are both closely related to the test sample and diverse enough to improve the model's robustness. 
\begin{figure}[t]
    \centering
        \includegraphics[width=.5\textwidth]{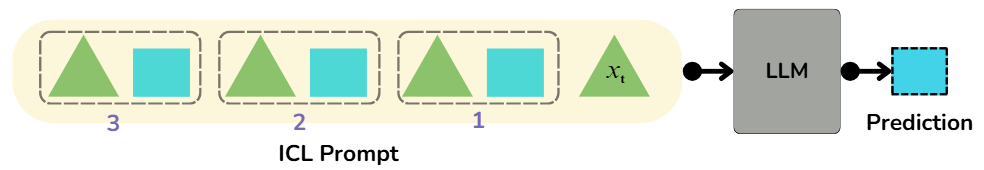} % Adjust the `trim` values as needed  
    \caption{Triangles represent SMILES strings, and squares are the labels. The ICL samples are appended in reverse order of retrieval.}% 
    \label{fig:prompt}
   % \vspace{-10pt}
\end{figure}
The selected demonstrations are appended in the prompt in reverse order as depicted in~\cref{fig:prompt}, which improves prediction compared to other permutations~\cite{lu-etal-2022-fantastically}.

%\suhang{add some sentences about how the demonstrations are used for ICL, maybe use a figure to help explain???}
\begin{table*}[t!]
   \centering
    \caption{Molecule captioning test results using different ICL retrieval methods}
    \label{tab:caption}
    \vskip -1em
   \begin{adjustbox}{width=.7\textwidth}
   \begin{tabular}{l|l|l|rrrrrrr}
   \toprule
   
   \multirow{2}{*}{\textbf{Dataset}} &
   \multirow{2}{*}{\textbf{Model}} & 
   \multirow{2}{*}{\textbf{Method}} & 
   \multicolumn{6}{c}{\textbf{Results}}  \\ 
   & & &
    BLEU-2 & BLEU-4 & ROUGE-1 & 
    ROUGE-2 & ROUGE-L & METEOR \\
   \midrule
   
   \multirow{12}{*}{\textbf{ChEBI-20}} &
   \multirow{4}{*}{\textbf{Mistral}}  
 & Random             &     0.229 & 0.125 & 0.325 & 0.152 & 0.273 & 0.287 \\         
 & & Scaffold         &     0.380 & 0.281 & 0.447 & 0.288 & 0.391 & 0.396 \\     
% & & KV-PLM         &0.531 & 0.424 & 0.612 & 0.456 & 0.552 & 0.577\\
 & & GAE &0.492 & 0.386 & 0.574 & 0.414 & 0.515 & 0.536\\
 & & \texttt{GAMIC}   &     \textbf{0.542} & \textbf{0.439} & \textbf{0.617} & \textbf{0.466} & \textbf{0.561} & \textbf{0.585} \\ 
    
   \cmidrule{2-9}
 & \multirow{4}{*}{\textbf{OpenChat}} 
 & Random             & 0.218 & 0.119 & 0.331 & 0.158 & 0.276 & 0.263 \\         
 & & Scaffold         & 0.363 & 0.269 & 0.446 & 0.286 & 0.391 & 0.381 \\     
 %& & KV-PLM         &0.518 & 0.415 & 0.608 & 0.453 & 0.550 & 0.566\\
  & & GAE &0.477 & 0.375 & 0.569 & 0.410 & 0.511 & 0.522\\
 & & \texttt{GAMIC}   & \textbf{0.527} & \textbf{0.427} & \textbf{0.612} & \textbf{0.462} & \textbf{0.558} & \textbf{0.571} \\ 
   \cmidrule{2-9}
 & \multirow{4}{*}{\textbf{Zephyr}} 
 & Random             &  0.177 & 0.093 & 0.304 & 0.139 & 0.258 & 0.252 \\        
 & & Scaffold         &  0.369 & 0.271 & 0.446 & 0.283 & 0.390 & 0.397 \\    
 %& & KV-PLM         &0.518 & 0.411 & 0.599 & 0.442 & 0.540 & 0.563\\
  & & GAE &0.477 & 0.372 & 0.561 & 0.401 & 0.503 & 0.521\\
 & & \texttt{GAMIC}   &  \textbf{0.526} & \textbf{0.422} & \textbf{0.605} & \textbf{0.451} & \textbf{0.548} & \textbf{0.570}\\
   \midrule
      \multirow{12}{*}{\textbf{PubChem}} &
   \multirow{4}{*}{\textbf{Mistral}}  
 & Random               & 0.155 & 0.084 & 0.251 & 0.122 & 0.215 & 0.210 \\
 & & Scaffold           & 0.261 & 0.182 & 0.371 & 0.229 & 0.323 & 0.343 \\
  & & GAE &0.318 & 0.242 & 0.437 & 0.299 & 0.390 & 0.403\\
 & & \texttt{GAMIC}     & \textbf{0.340} & \textbf{0.262} & \textbf{0.455} & \textbf{0.317} & \textbf{0.407} & \textbf{0.421} \\
   \cmidrule{2-9}
 & \multirow{4}{*}{\textbf{OpenChat}} 
 & Random               & 0.128 & 0.067 & 0.251 & 0.119 & 0.212 & 0.215 \\
 & & Scaffold           & 0.203 & 0.140 & 0.360 & 0.221 & 0.313 & 0.336 \\
  & & GAE &0.302 & 0.226 & 0.428 & 0.289 & 0.381 & 0.395\\
 & & \texttt{GAMIC}     & \textbf{0.311} & \textbf{0.236} & \textbf{0.443} & \textbf{0.305} & \textbf{0.396} & \textbf{0.413} \\
   \cmidrule{2-9}
 & \multirow{4}{*}{\textbf{Zephyr}} 
 & Random               & 0.149 & 0.080 & 0.250 & 0.121 & 0.214 & 0.206 \\
 & & Scaffold           & 0.262 & 0.180 & 0.367 & 0.220 & 0.316 & 0.326 \\
  & & GAE &0.310 & 0.235 & 0.427 & 0.291 & 0.382 & 0.392\\
 & & \texttt{GAMIC}     & \textbf{0.323} & \textbf{0.246} & \textbf{0.441} & \textbf{0.304} & \textbf{0.394} & \textbf{0.406} \\

 %& & KV-PLM 
   \bottomrule
   \end{tabular}
   \end{adjustbox}
\end{table*}
\section{Experiments}
In this section, we conduct experiments to verify the effectiveness of the proposed framework. In particular, we aim to answer the following research questions: %How does the performance of ICL with \texttt{GAMIC} compare to conventional methods in: 
(\textbf{RQ1}) \textit{Molecular Performance Analysis}: How does the performance of ICL with \texttt{GAMIC} compare to other ICL methods for various classes of molecule analysis tasks?
(\textbf{RQ2}) \textit{Sensitivity Analysis}: How sensitive is \texttt{GAMIC} w.r.t to the number of demonstrations?
(\textbf{RQ3}) \textit{Ablation Study}: How does each element contribute to \texttt{GAMIC}?
%\suhang{I prefer research questions to be:  Q1: How does the performance of ICL with \texttt{GAMIC} compare to other ICL methods for various kinds of molecule analysis tasks? Q2: How sensitive is \texttt{GAMIC} w.r.t to the number of demonstrations? Q3: How does each component contribute to \texttt{GAMIC}?}

%How \textbf{RQ1}: How does the performance of in-context learning with \texttt{GAMIC} compare to conventional methods in various molecular tasks? \textbf{RQ2}: How does the performance of in-context learning with our graph-based contrastive sample selection compare to conventional methods in various molecular tasks? \zhiwei{Ali, could you come up a RQ3? It is better to have 3 RQ}

\subsection{Experiment Setup}

\subsubsection{Datasets} We evaluate our approach on three representative molecular tasks:  molecule captioning, molecule property prediction, and molecule yield prediction, which represent three different molecular task classes (See Table \ref{tab:dataset_stats}).  For each task, we utilize two or more datasets as follows:
\begin{itemize}[leftmargin=*]
    \item \textbf{Molecule Captioning}: We evaluate performance on molecule captioning using the test split of ChEBI-20~\cite{edwards2021text2mol}. This dataset provides a focused assessment of bidirectional translation between molecular structures and natural language descriptions. \textit{We also utilize the training set of %\suhang{the training set of} 
    this dataset to train \texttt{GAMIC}}. In addition, we utilize the split suggested by Liu et al.~\cite{liu2022multimodal} to evaluate the PubChem~\cite{pubchem} dataset.
    \item \textbf{Property Prediction}: Datasets \textit{BBBP}, \textit{BACE}, \textit{HIV}, \textit{Tox21}, and \textit{ClinTox} proposed by~\cite{wu2018moleculenet} are binary classification datasets that consist of SMILES strings, and binary labels of specific molecular properties, which we use to assess the accuracy of the predictions.
    \item \textbf{Yield Prediction}: We utilize Suzuki-Miyaura~\cite{reizman2016suzuki}, and Buchwald-Hartwig~\cite{ahneman2018predicting} datasets which include molecule reactions and their corresponding yield which can be classified as high or low. 
\end{itemize}
For datasets without a predefined test split, we create three random train-valid-test splits by 8:1:1 ratio, following standard practice in the literature~\cite{wang2021chemical} using predefined random seeds. We conduct experiments on each split and report the average results across the three runs. Table~\ref{tab:dataset_stats} summarizes the key statistics of the datasets.

\subsubsection{Baselines Molecular ICL Methods}
As our framework focuses on ICL, we compare \texttt{GAMIC} with representative and state-of-the-art ICL methods for molecular analysis, including: (1) \textbf{Random Selection}, which selects samples for the demonstration pool at random without replacement; %\suhang{which xxx} 
(2) \textbf{Scaffold}~\cite{guo_what_2023}, which utilizes Tanimoto similarity~\cite{bajusz2015tanimoto} between the Morgan fingerprints of the test sample and the demonstrations to return the top $k$ demonstrations. The demonstrations are appended in reverse order as in~\cref{fig:prompt}; and (3) \textbf{GAE}, which utilizes graph autoencoder~\cite{kipf2016variational} to learn graph representations. Specifically, it adopts a two-layer GAT followed by a pooling layer to obtain graph representation for a molecular graph, then reconstruct the adjacency matrix with an MLP and adopts mean square loss between the original adjacency matrix and the reconstructed adjacency matrix as the loss function to train the autoencoder. Once the model is trained, the encoder can utilize latent structure for retrieving similar molecules.
%\suhang{GAT will not give the graph representation, you need the pooling layer.} The decoder is defined as follows:
%\[
%\hat{\mathbf{x}} = \text{MLP}(\mathbf{h}; \mathbf{w}^{(1)})
%\]
%\suhang{what is $\mathbf{x}$, shouldn't autoencoder reconstruct the adjacency matrix???}  A standard MSE reconstruction loss is used:
%where \(N\) is the number of nodes in the graph. During retrieval, we use $\text{MeanPool(h)}$ on encoded graphs and retrieve top $k$ similar samples.
%\suhang{For baslines, we usually use plain text to explain, e.g., GAE, which utilizes graph author encoder (cite) to learn graph representations. Specifically, it adopts a two layer GAT followed by a pooling layer to obtain graph representation for a molecualr graph, then reconstruct the adjacency matrix with an MLP and adopts mean square loss between xxx and xxx as the loss function to train the authorencoder. Once the model is trained, the encoder can xxx for retrieving xxx}

%~\suhang{cite}, \suhang{which xxx (explain it)}. On molecule captioning with ChEBI-20, we will also perform an ablation test to measure the contribution of different design choices on our outcome.%also assess our method against a pretrain-finetune method, namely MolT5.
\begin{table*}[!t]
\centering
\caption{Property Prediction F1-score and a summarized mean score}%\suhang{why for Mistral with Random, the performance on HIV is 0???\ali{I have run it multiple times, but can't get any True Positives}}}%\ali{w/o mmr will be removed}}
\label{tab:prop}
\vskip -1em
\begin{adjustbox}{width=.9\textwidth}
\begin{tabular}{@{\extracolsep{\fill}}p{1.8cm}l|rrrrr|r}
\toprule
Model & Method & BBBP & BACE & HIV & Tox21 & ClinTox & {\footnotesize All Data Mean} \\
\midrule
\multirow{3}{*}{Mistral}
    & Random        & 0.694 $\pm$ 0.032&0.372 $\pm $0.062&0  & 0.037 $\pm$ 0.025 & 0.011 $\pm$ 0.043&0.223\\
    & Scaffold      & 0.850 $\pm$ 0.494&0.710 $\pm$ 0.093&0.392 $\pm$ 0.216 & 0.203 $\pm$ 0.099 & 0.100 $\pm$ 0.087&0.451\\
    & GAE & 0.858 $\pm$ 0.012&0.701 $\pm$ 0.053&0.289 $\pm$ 0.012 & 0.216 $\pm$ 0.068 & 0.103 $\pm$ 0.178&0.433 \\
    & \texttt{GAMIC}&\textbf{0.905} $\pm$ 0.031&\textbf{0.726} $\pm$ 0.127&\textbf{0.400} $\pm$ 0.202&\textbf{0.271} $\pm$ 0.064&\textbf{0.112} $\pm$ 0.040&\textbf{0.483}\\
    %&W/o MMR &0.892 $\pm$ 0.018&\textbf{0.728} $\pm$ 0.007&\textbf{0.434} $\pm$ 0.005&\textbf{0.286} $\pm$ 0.016&\textbf{0.122} $\pm$ 0.107&0.493\\
\midrule
\multirow{3}{*}{OpenChat} 
    & Random         & 0.289 $\pm$ 0.051&0.525 $\pm$ 0.005&0.012 $\pm$ 0.013 & 0.008 $\pm$ 0.013 & 0.044 $\pm$ 0.077&0.176 \\
    & Scaffold       & 0.749 $\pm$ 0.022&0.665 $\pm$ 0.053&0.364 $\pm$ 0.018 & 0.111 $\pm$ 0.085 & 0.083 $\pm$ 0.144&0.394\\
    & GAE & 0.745 $\pm$ 0.013&0.674 $\pm$ 0.021&0.315 $\pm$ 0.055 & 0.131 $\pm$ 0.059 & 0.048 $\pm$ 0.082&0.383\\
    & \texttt{GAMIC} &\textbf{0.836} $\pm$ 0.024&\textbf{0.674} $\pm$ 0.037&\textbf{0.365} $\pm$ 0.019&\textbf{0.153} $\pm$ 0.019&\textbf{0.203} $\pm$ 0.093&\textbf{0.446}  \\
    %&W/o MMR &0.839 $\pm$ 0.018&\textbf{0.655} $\pm$ 0.035&\textbf{0.408} $\pm$ 0.003&\textbf{0.196} $\pm$ 0.042&\textbf{0.149} $\pm$ 0.136&0.450\\
\midrule
\multirow{3}{*}{Zephyr} 
    & Random        & 0.518 $\pm$ 0.034&0.750 $\pm$ 0.032&0.020 $\pm$ 0.009 & 0.095 $\pm$ 0.040 & 0.139 $\pm$ 0.127&0.304 \\
    & Scaffold      & 0.875 $\pm$ 0.004&0.769 $\pm$ 0.040&0.386 $\pm$ 0.054 & 0.242 $\pm$ 0.046 & 0.242 $\pm$ 0.162&0.503\\
    & GAE & 0.881 $\pm$ 0.022&0.747 $\pm$ 0.065&0.326 $\pm$ 0.037 & 0.246 $\pm$ 0.021 & 0.169 $\pm$ 0.177&0.474\\
    & \texttt{GAMIC}&\textbf{0.924} $\pm$ 0.009&\textbf{0.783} $\pm$ 0.034&\textbf{0.422} $\pm$ 0.011&\textbf{0.276} $\pm$ 0.023&\textbf{0.361} $\pm$ 0.127&\textbf{0.553}  \\
    %&W/o MMR &0.917 $\pm$ 0.012&\textbf{0.763} $\pm$ 0.049&\textbf{0.419} $\pm$ 0.030&\textbf{0.283} $\pm$ 0.025&\textbf{0.390} $\pm$ 0.16&0.555\\
\bottomrule
\end{tabular}
\end{adjustbox}
\end{table*}

\subsubsection{LLM Models}
%For \textbf{RQ1}, we evaluate GPT-4's performance against existing SOTA benchmark tasks results reported across our diverse and complex datasets: PubChem, ChemLLMBench, and ChEBI-20 benchmarks. This allows us to establish a strong baseline for comparison with current leading models. For \textbf{RQ2}, 
To show that our \texttt{GAMIC} is flexible to facilitate various LLM backbones, we conduct comprehensive evaluations using three representative small to medium-sized Language Models (LLMs), selected for their diversity in architecture and training approaches, which include \begin{inparaenum}[(1)] \item \textbf{Mistral-7B}~\cite{jiang2023mistral}: A state-of-the-art model with 7 billion parameters, showcasing cutting-edge performance; \item \textbf{OpenChat-8B}~\cite{wang2023openchat}: An open-source conversational AI model, highlighting the strengths of publicly accessible systems; \item \textbf{Zephyr-7B}~\cite{tunstall2023zephyr}: A fine-tuned variant of the Mistral architecture, optimized for specialized tasks. \end{inparaenum}

\subsubsection{Evaluation Metrics}
For property prediction and yield prediction, we report the F1-score and the standard deviation. For molecule captioning, we employ a comprehensive set of text generation metrics used in the literature~\cite{guo_what_2023,li_empowering_2024} to evaluate molecular description quality: BLEU (BLEU-2, and BLEU-4), ROUGE (ROUGE-1, ROUGE-2, ROUGE-L), and METEOR. All metrics range from 0 to 1, with higher scores indicating better alignment between generated and reference molecular descriptions.

\subsubsection{Evaluation Setup}
For each task, we follow the benchmark's standard evaluation protocol by evaluating the test set, and utilizing the training set as a demonstration pool from which samples can be retrieved, as described in Table \ref{tab:dataset_stats}.

To account for the stochastic nature of LLM outputs, we perform five repeated evaluations for each experiment and report the mean of the results.
We evaluate our proposed method on the 9 different benchmark datasets across three molecular tasks. 

% For all datasets, we evaluate \texttt{GAMIC}'s effectiveness through comprehensive comparisons with \begin{inparaenum}[(a)]
%     \item Random selection, which serves as a basic baseline,
%     \item Scaffold: Morgan fingerprint-based similarity metrics, and
%     \item GAE: a graph similarity baseline.
% \end{inparaenum}

For molecule captioning, we use $k=2$ to control the prompt length as the labels for this task are long textual descriptions. For other tasks, we use $k=3$. In addition, for all experiments, we use $\lambda = 0.3$.
%Unless otherwise stated, all experiments utilize two demonstration samples ($k$=2).
%\suhang{how about $\lambda$?}

\begin{figure}[t]
    \centering
        \includegraphics[width=.51\textwidth]{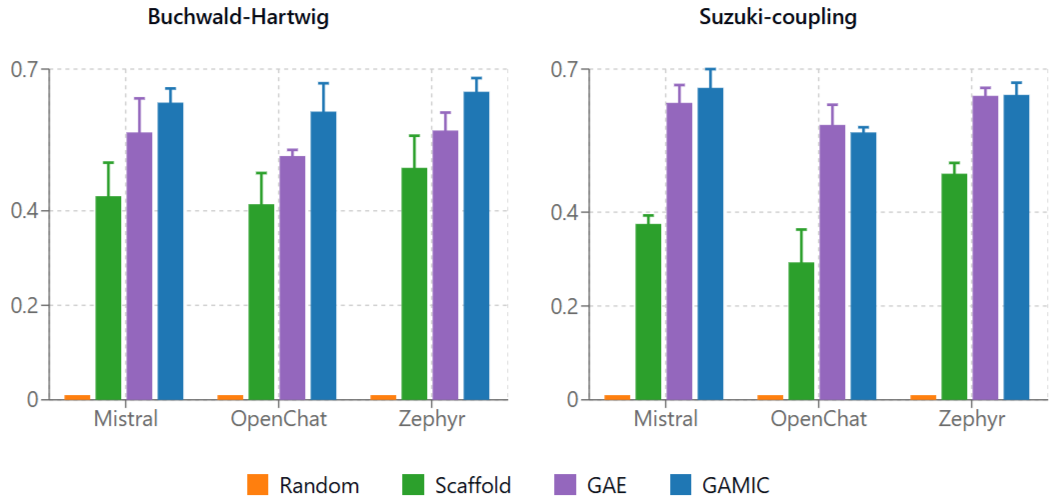} % Adjust the `trim` values as needed  
    \caption{Yield prediction F1-score}%\suhang{the solution of the figures are low }}% \suhang{this figure looks good. I think we might need to have another figure showing the ICL process with MMR}}
    \label{fig:yield}
   % \vspace{-10pt}
\end{figure}
\begin{figure}[t]
    \centering
        \includegraphics[width=.4\textwidth]{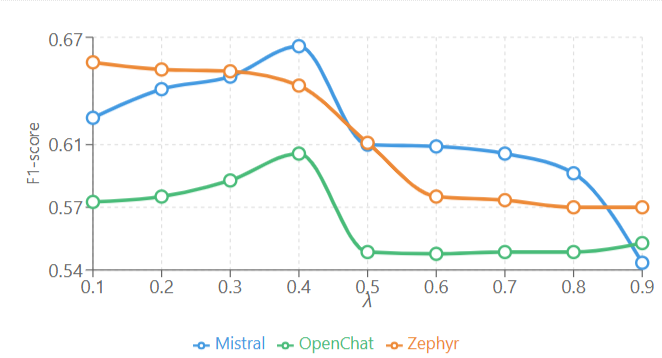} % Adjust the `trim` values as needed  
    \caption{$\lambda$ sensitivity analysis using average Yield prediction} %\suhang{consider using line plot. we have too many bar plots} \suhang{the plot is a little bit strange to me. Usually, we should observe a pattern that the performance first increase then decrease. However, in this figure when $\lambda = 0.5$, the performance is the worst? Could you double check if there's anything wrong with the experiments or when you copy the results? Otherwise, can you explain why we have such observation? Reviewers will ask about this}}% \suhang{this figure looks good. I think we might need to have another figure showing the ICL process with MMR}}
    \label{fig:lambda}
   % \vspace{-10pt}
\end{figure}

\begin{table*}[tbh!]
   \centering
   \caption{\texttt{GAMIC} ablation results on molecule captioning using ChEBI-20 dataset} %\suhang{the results just shows that MMR is not important. This will be criticized by reviewers as you described MMR as one contribution. How about some other tasks, can we see some improvement with MMR? I suggest to replace this table with some tasks that can show the importance of each component.}}
   \label{tab:caption-ab}
   \vskip -1em
   \begin{adjustbox}{width=.9\textwidth}

   \begin{tabular}{l|l|cc|rrrrrrr}
   \toprule
   \multirow{2}{*}{\small{Model}} & 
   \multirow{2}{*}{\small{Method}} & 
   \multirow{2}{*}{\small{Morgan S}} & 
   \multirow{2}{*}{\small{SciBERT}} & 
   \multicolumn{6}{c}{\small{Results}}  \\ 
   & & & &
    \small{BLEU-2} & \small{BLEU-4} & \small{ROUGE-1} & 
    \small{ROUGE-2} & \small{ROUGE-L} & \small{METEOR} \\
   \midrule
   \multirow{4}{*}{\textbf{Mistral}} 
        & \small{W/o Morgan-BERT} &\xmark & \xmark  & 0.520 & 0.415 & 0.599 & 0.444 & 0.541 & 0.566\\
        & \texttt{GAMIC}\small{-BERT} &\cmark & \xmark               &0.533 & 0.430 & 0.611 & 0.457 & 0.553 & 0.577\\
        & \small{W/o Morgan} &\xmark & \cmark               &0.535 & 0.431 & 0.613 & 0.460 & 0.554 & 0.580\\
        & \texttt{GAMIC} & \cmark & \cmark          &\textbf{0.542} & \textbf{0.439} & \textbf{0.617} & \textbf{0.466} & \textbf{0.561} & \textbf{0.585}\\
   \midrule   
   \multirow{4}{*}{\textbf{OpenChat}}  
        & \small{W/o Morgan-BERT} &\xmark & \xmark  &0.505 & 0.404 & 0.594 & 0.441 & 0.538 & 0.551\\
        & \texttt{GAMIC}\small{-BERT} &\cmark & \xmark               &0.518 & 0.418 & 0.604 & 0.452 & 0.548 & 0.562\\
        & \small{W/o Morgan} &\xmark & \cmark               &0.522 & 0.421 & 0.608 & 0.456 & 0.552 & 0.566\\
        & \texttt{GAMIC} & \cmark & \cmark          &\textbf{0.527} & \textbf{0.427} & \textbf{0.613} & \textbf{0.462} & \textbf{0.557} & \textbf{0.571}\\
   \midrule

   \multirow{4}{*}{\textbf{Zephyr}} 
        & \small{W/o Morgan-BERT} &\xmark & \xmark  &0.508 & 0.404 & 0.589 & 0.434 & 0.532 & 0.553\\
        & \texttt{GAMIC}\small{-BERT} &\cmark & \xmark               &0.520 & 0.416 & 0.600 & 0.445 & 0.543 & 0.565\\
        & \small{W/o Morgan} &\xmark & \cmark               &0.521 & 0.416 & 0.602 & 0.447 & 0.545 & 0.567\\
        & \texttt{GAMIC} & \cmark & \cmark          &\textbf{0.526} & \textbf{0.422} & \textbf{0.605} & \textbf{0.451} & \textbf{0.548} & \textbf{0.570}\\
   \bottomrule
   \end{tabular}
   \end{adjustbox}
\end{table*}

\begin{table}[h!]
   \centering
    \caption{Sensitivity analysis for different ICL demonstration sample sizes ($k$) on molecule captioning} \label{tab:sensitivity}
    \vskip -1em
\begin{adjustbox}{width=.45\textwidth}
   \begin{tabular}{l|c|rrrrrrr}
   \toprule
   \multirow{2}{*}{\textbf{Model}} & 
   \multirow{2}{*}{$k$} & 
   \multicolumn{6}{c}{\textbf{Results}}  \\ 
   & &
    \scriptsize{BLEU-2} & 
    \scriptsize{BLEU-4} & 
    \scriptsize{ROUGE-1} & 
    \scriptsize{ROUGE-2} & 
    \scriptsize{ROUGE-L} & 
    \scriptsize{METEOR} \\
   \midrule
   
   \multirow{6}{*}{\textbf{Mistral}}  
 & 0        &0.055 & 0.023 & 0.135 & 0.065 & 0.123 & 0.073\\
 & 1 &0.536 & 0.431 & 0.612 & 0.459 & 0.554 & 0.581\\
 & 2 &0.542 & 0.439 & 0.617 & 0.466 & 0.561 & 0.585\\
 & 3 &\textbf{0.543} & \textbf{0.440} & \textbf{0.619} & \textbf{0.468} & \textbf{0.563} & \textbf{0.586}\\
 & 4 &0.531 & 0.426 & 0.609 & 0.454 & 0.551 & 0.573 \\
 & 5&0.530 & 0.425 & 0.609 & 0.454 & 0.551 & 0.573\\
 &10&0.528 & 0.423 & 0.605 & 0.450 & 0.547 & 0.572\\
   \midrule
   \multirow{6}{*}{\textbf{OpenChat}} 
 & 0 &0.037 & 0.007 & 0.101 & 0.011 & 0.083 & 0.067\\
 & 1 &0.523 & 0.422 & 0.606 & 0.455 & 0.550 & 0.569\\
 & 2 &0.527 & 0.427 & 0.613 & 0.462 & 0.557 & 0.571\\
 & 3 &\textbf{0.528} & 0.427 & \textbf{0.614} & \textbf{0.461} & \textbf{0.557} & \textbf{0.573}\\ 
 & 4 &0.518 & 0.416 & 0.603 & 0.449 & 0.547 & 0.563 \\
  & 5&0.521 & 0.419 & 0.609 & 0.456 & 0.553 & 0.569\\
 &10&0.518 & 0.415 & 0.605 & 0.449 & 0.549 & 0.563\\
   \midrule
   \multirow{6}{*}{\textbf{Zephyr}} 
 & 0 &0.048 & 0.005 & 0.130 & 0.018 & 0.100 & 0.082\\
 & 1 &0.514 & 0.409 & 0.592 & 0.438 & 0.535 & 0.558\\
 & 2 &0.526 & 0.422 & 0.605 & 0.451 & 0.548 & 0.570\\
 & 3 &0.526 & \textbf{0.423} & \textbf{0.609} & \textbf{0.455} & \textbf{0.552} & 0.570\\
 & 4 &0.524 & 0.419 & 0.606 & 0.451 & 0.549 & 0.568 \\
  & 5&0.520 & 0.416 & 0.605 & 0.449 & 0.547 & 0.565\\
 &10&0.518 & 0.412 & 0.599 & 0.442 & 0.540 & 0.563\\
   \bottomrule
   \end{tabular}
\end{adjustbox}
\end{table}

% \paragraph{\textbf{RQ2}:} How does the performance of in-context learning with our graph-based contrastive sample selection compare to conventional methods in various molecular tasks? 

%we assess \texttt{GAMIC}'s ability to enhance general-purpose LLMs on molecule captioning, which evaluates models' ability to describe molecular structures
%\end{inparaenum}. We use established benchmarks including ChEBI-20 and PubChem.
\subsection{RQ1. Molecular Performance Analysis}
%\suhang{Can we change the order of presenting the results, putting and presenting better results/tasks first? }
%\suhang{I think we can add a little bit more analysis of the experimental results to show the effectiveness of the proposed framework}

\noindent\textbf{Molecule Explaining}. \Cref{tab:caption} presents the results of \texttt{GAMIC} compared to benchmark methods on ChEBI-20 and PubChem datasets. \texttt{GAMIC} significantly outperforms other models across all evaluation metrics. This validates that graph representations capture the complex relationships of molecules more effectively. Furthermore, this demonstrated the effectiveness of \texttt{GAMIC} in overcoming the modality gap and dataset limitations present in both datasets.

\noindent\textbf{Molecular Reasoning}. 
As \cref{fig:yield} shows, \texttt{GAMIC} significantly improves the accuracy of yield prediction across all dataset/LLM combinations, which demonstrates it's effectiveness in overcoming the GNN complexity challenge. Hence, chemical validity is preserved in yield prediction more effectively than other baseline methods.

Moreover, random selection performs extremely poorly on both datasets on this task. On the other hand, GAE outperforms Scaffold, which validates the importance of graphs in effectively representing molecules. %This underlines the relative equivalence of these LLMs in molecular reasoning. 
%Conversely, the improvement gains of Mistral underscore its capacity to make the best use of ICL demonstrations compared to OpenChat, and Zephyr.

\noindent\textbf{Molecular Understanding}. %\suhang{can you also put results without demonstration?} \ali{Unable to get correct Yes/No responses without ICL for all LLMs}}
\Cref{tab:prop} shows the results for molecular understanding. \texttt{GAMIC} provides the best overall results on average, while Scaffold outperforms random selection. On the HIV dataset using Random retrieval, Mistral reports an F1-score of 0, indicating a failure to achieve any True Positives.

Overall, \texttt{GAMIC} outperforms the baselines on all property prediction benchmarks. The effectiveness of \texttt{GAMIC} on this task further corroborates its capacity to preserve chemical validity in cross-modal training.

% In several tests where Scaffold outperforms \texttt{GAMIC}, the results are insignificant, particularly in the Tox21, and ClinTox datasets. Furthermore, ClinTox reports high accuracy with Random selection, which leaves little room for improvement with advanced retrieval mechanisms, such as ours. On average, the results highlight the effectiveness of \texttt{GAMIC} in molecular understanding.

\subsection{RQ2: Sensitivity Analysis} 
We conduct a sensitivity analysis to assess how the molecule captioning performs in response to additional demonstration samples. Specifically, we vary the number of demonstrations as $\{0,1,2,3,5,10\}$ and the results are given in Table~\ref{tab:sensitivity}. The results plateau at three ICL samples and there is insignificant improvement between $k=2$, and $k=3$, which further motivates our selection of $k=2$ for this task to control prompt length. As we increase $k>3$, the performance begins to deteriorate slowly.%suhang{add analysis of the experimental results. Please also see my comments in the table}

Furthermore, we analyze how modifying the MMR parameter, $\lambda$, affects the prediction outcome. We fix $k$ as 3 and vary $\lambda$ from 0.1 to 0.9. The results are shown in \Cref{fig:lambda}. Based on the figure, we can observe that values of 0.3 or 0.4 appear plausible choices. 

%\suhang{there is another hyperparameter, i.e., $\lambda$. Do you have hyperparameter sensitivity analysis for $\lambda$??}

\subsection{RQ3: Ablation Study}

We conduct a focused ablation study to evaluate the contribution of each module to our framework by comparing it against the following variants: \begin{inparaenum}[(i)] \item \label{i} W/o Morgan-BERT: During training, this method uses only the corresponding caption as the positive pair, and other samples as negative pairs. It also encodes captions with BERT, which has a limited scientific vocabulary, rather than SciBERT. This helps isolate the contributions of SciBERT and Morgan sampling; \item \texttt{GAMIC}-BERT: Uses Morgan sampling during training, but encodes captions with BERT instead of SciBERT; 
\item W/o Morgan: Similar to (\ref{i}), but encodes captions using SciBERT to quantify the contribution of SciBERT. \end{inparaenum}

\Cref{tab:caption-ab} demonstrates the contribution of Morgan sampling and SciBERT compared to W/o Morgan-BERT. Both approaches contribute similarly to individual improvements, with a slight advantage for using SciBERT. The combined contribution of both elements leads to better performance than either method alone.

\begin{figure}[t]
    \centering
        \includegraphics[width=.5\textwidth]{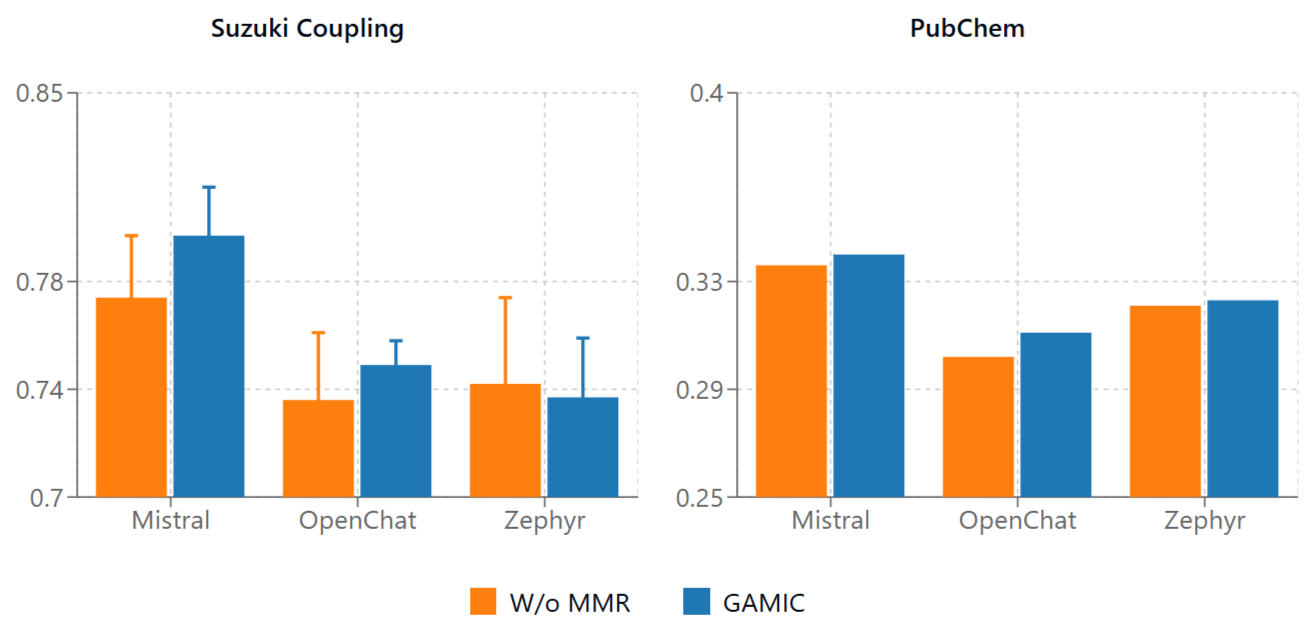} % Adjust the `trim` values as needed  
    \caption{MMR vs W/o MMR on Suzuki dataset accuracy (left) and PubChem BLEU score (right)}% \suhang{this figure looks good. I think we might need to have another figure showing the ICL process with MMR}}
    \label{fig:mmr-ab}
   % \vspace{-10pt}
\end{figure}

Additionally, we evaluate the contribution of MMR by comparing it with W/o MMR, which retrieves the top $k$ most similar samples, ordered in reverse similarity, as shown in \Cref{fig:mmr}. \Cref{fig:mmr-ab} illustrates the improvement of MMR in yield and property prediction averages. It shows that MMR provides better results across multiple tasks and for all LLMs tested.

%\suhang{Could you first describe the variants, many details are missing, e.g., (i) W/o MMR: To evaluate xxx, we design xxx, which xxx; (ii) W/o Morgan: (if without Morgan, how do you prepare the positive and negative); (iii) W/o SciBERT: (W/o SciBERT, what do we use for the text part, or it is without the text part????)} %The ablation experiments are confined to the molecule captioning task due to its broader range of evaluation metrics, providing a more comprehensive assessment of performance variations. 

%\Cref{tab:caption-ab} demonstrates the impact of our approach in isolation, i.e. using BERT instead of SciBERT, and when using SciBERT. 

%This may be related to the sensitivity analysis below (see \Cref{tab:sensitivity}) which demonstrates that performance may decline with more demonstration examples, which is in line with results reported by Guo, et al~\cite{guo_what_2023}. 

%\suhang{we have too many tables. Consider replacing some tables with figures}

%\section{Discussion}

\section{Conclusions}
This work demonstrates the potential of medium-sized Large Language Models (LLMs) in molecular understanding. We focus on smaller LLMs (7–10B parameters) due to their lower computational costs and ease of deployment in real-world applications. Our results demonstrate the capacity of these LLMs to perform multiple molecular tasks without task-specific fine-tuning using advanced demonstration selection techniques. We introduced \texttt{GAMIC}, which achieves state-of-the-art performance in molecular ICL. These findings bridge the gap between molecular structure representation and LLM capabilities, advancing applications in drug discovery and materials science.

%As future work, we are interested in investigating the capacity of molecule design at a deeper level. It would be highly important for researchers in the chemical domain to alter molecule properties using regular text, and see the effect of these changes on the molecular graph, thereby creating specialized molecules with the given features. This may require inventing new approaches for molecule design, but also specific benchmark datasets to advance this work.

\appendix

\bibliographystyle{named}
\bibliography{MAIN}
\appendix
\section{Case Study}
\begin{figure*}[ht]
    \centering
        \includegraphics[width=1\textwidth]{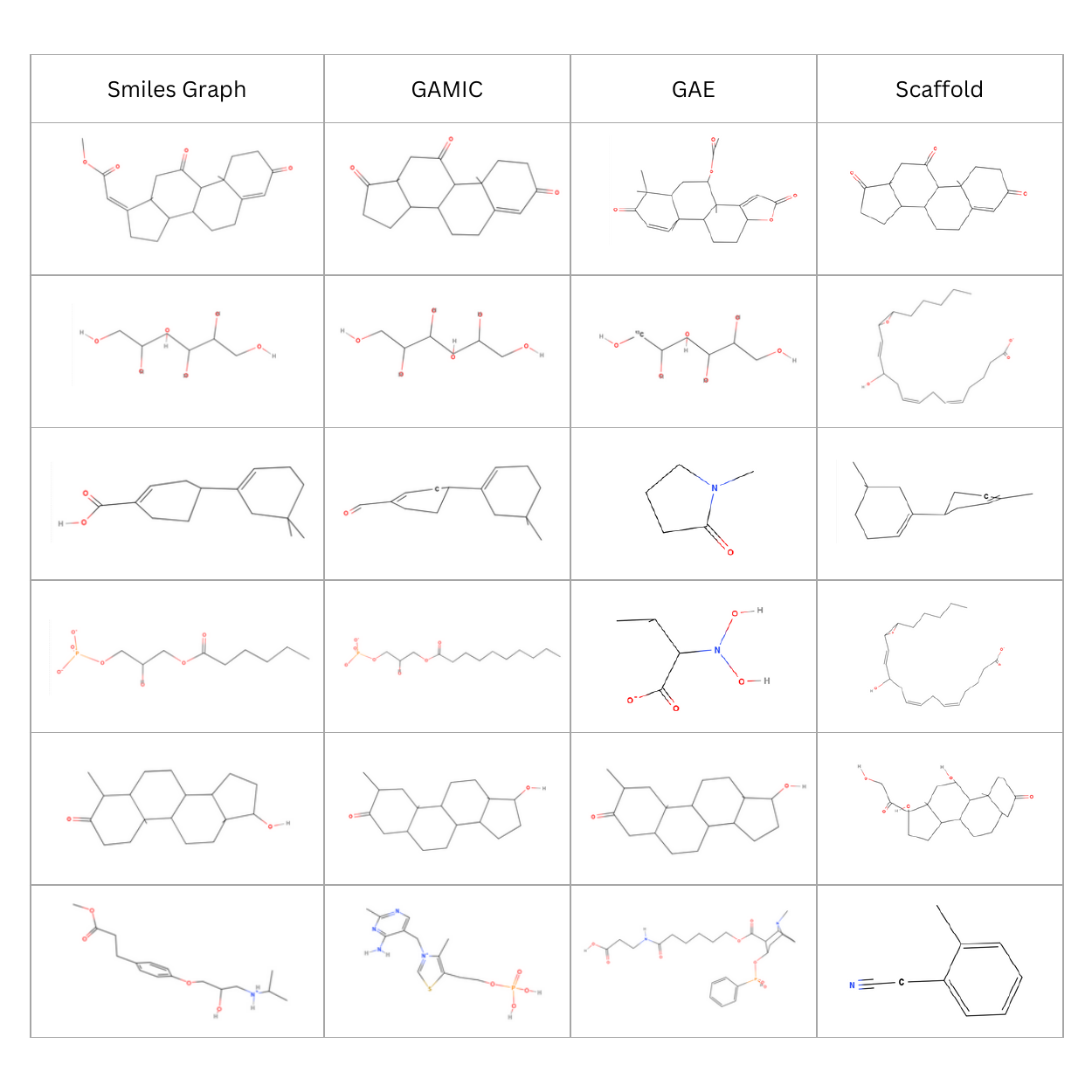} % Adjust the `trim` values as needed  
    \caption{Retrieval examples using various methods}%\suhang{the solution of the figures are low }}% \suhang{this figure looks good. I think we might need to have another figure showing the ICL process with MMR}}
    \label{fig:case}
   % \vspace{-10pt}
\end{figure*}

\Cref{fig:case} illustrates retrieved molecules for a set of test molecule using \texttt{GAMIC}, and other baselines. 

\section{Prompt for zero-shot Molecule Captioning}
For zero-shot molecular captioning results, we utilize the following prompt:
\begin{tcolorbox}[colback=blue!10, colframe=blue!40!black, title=Zero-shot Prompt, rounded corners]
You are an expert chemist. Given the molecular SMILES, your task is to predict the molecule description using your experienced molecular knowledge.

SMILES:[SMILE String]

Caption:
\label{prompt:GPT-4}
\end{tcolorbox}

For multi-shot, we do not include instructions. Instead, we directly put the demonstrations in input/output format.

\section{Additional Data on Evaluation Metrics}
For molecular explanation we utilize the following metrics:
\begin{itemize}
    \item \textbf{BLEU} (Bilingual Evaluation Understudy)~\cite{papineni2002bleu}: We use BLEU-2 and BLEU-4 scores to assess n-gram precision between generated and reference texts. BLEU-2 captures local phrase matching, while BLEU-4 evaluates longer sequence accuracy.
    
   \item \textbf{ROUGE} (Recall-Oriented Understudy for Gisting Evaluation)~\cite{lin2004rouge}: We utilize three variants: \begin{inparaenum}[(1)]
            \item ROUGE-1: Measures unigram overlap
            \item ROUGE-2: Assesses bigram overlap
            \item ROUGE-L: Evaluates longest common subsequence, capturing flexible sequence matching
       \end{inparaenum}
        
    \item \textbf{METEOR} (Metric for Evaluation of Translation with Explicit ORdering)~\cite{banerjee2005meteor}: Provides a more nuanced evaluation by incorporating synonyms, stemming, and word order, better capturing semantic similarity.
\end{itemize}

\end{document}